\documentclass[runningheads]{llncs}
\usepackage[compatibility=false]{caption}
\usepackage[T1]{fontenc}
\usepackage{graphicx}
\usepackage{amsmath,amssymb,amsfonts}
\usepackage{textcomp}
\usepackage{xcolor}
\usepackage{url}
\usepackage{booktabs}
\usepackage{multirow}
\usepackage{array}
\usepackage{tabularx}
\usepackage{longtable}
\usepackage{comment}
\usepackage{parskip}
\usepackage{float}
\usepackage{caption}
\usepackage{subcaption}
\usepackage{hyperref}
\makeatletter
\renewcommand{\@listi}{%
  \leftmargin\leftmargini
  \topsep 2pt
  \parsep 0pt
  \itemsep 2pt}
\makeatother

\begin{document}

\title{BdSL-SPOTER: A Transformer-Based Framework for Bengali Sign Language Recognition with Cultural Adaptation}

\author{Sayad Ibna Azad\inst{1}\orcidID{0009-0006-2301-9160} \and 
Md. Atiqur Rahman\inst{1}\orcidID{0009-0000-2322-6051}}

\institute{Department of Computer Science and Engineering \\
Islamic University of Technology \\
Board Bazar, Gazipur-1704, Bangladesh \\
\email{sayadibnaazad@iut-dhaka.edu} \and 
\email{atiqurrahman23@iut-dhaka.edu}}
\maketitle

\begin{abstract}
We introduce BdSL-SPOTER, a pose-based transformer framework for accurate and efficient recognition of Bengali Sign Language (BdSL). BdSL-SPOTER extends the SPOTER paradigm with cultural specific preprocessing and a compact four-layer transformer encoder featuring optimized learnable positional encodings, while employing curriculum learning to enhance generalization on limited data and accelerate convergence. On the BdSLW60 benchmark, it achieves 97.92\% Top-1 validation accuracy, representing a 22.82\% improvement over the Bi-LSTM baseline, all while keeping computational costs low. With its reduced number of parameters, lower FLOPs, and higher FPS, BdSL-SPOTER provides a practical framework for real-world accessibility applications and serves as a scalable model for other low-resource regional sign languages.
\end{abstract}


\section{Introduction}
According to current statistics, 70 million deaf individuals~\cite{world2021world} rely on visual-gestural language for daily interaction all over the world. These hearing-impaired communities use sign language as their primary communication medium. In Bangladesh, the number of hearing-impaired people is over 13.7 million, who use Bengali Sign Language (BdSL) as their main communication mode. However, there exists a significant communication gap between hearing-impaired individuals and the general population.

To bridge this communication gap, an emerging technological solution is the Automatic Sign Language Recognition (SLR) system. Recent advances in deep learning have improved sequence modeling tasks across various domains. The introduction of SPOTER, a pose-based transformer, demonstrated the potential of applying transformer architectures to sign language recognition using features extracted from MediaPipe~\cite{bohavcek2022sign}.

There has been significant progress in American Sign Language (ASL) and European Sign Language recognition, but research in Bengali Sign Language is still severely limited. Current benchmarks on the BdSLW60 dataset~\cite{rubaiyeat2025bdslw60}, containing 9,307 videos across 60 classes with 18 signers, show a maximum accuracy of only 75.1\% using Bi-LSTM architectures. This paper addresses these limitations by presenting BdSL-SPOTER, a novel transformer-based architecture that incorporates comprehensive cultural and architectural innovations, which fundamentally rethink transformer design. We have combined systematic architectural optimization and cultural adaptation techniques to achieve higher performance on limited training data.

The key contributions of this work includes,
\begin{itemize}
    \item Cultural adapted preprocessing pipeline tailored to Bengali signing space characteristics, temporal dynamics, and gesture patterns.
    \item Optimized transformer architecture featuring only four encoder layers and learnable positional encoding for efficient modeling.
    \item Effective training strategy requiring only 4.8 minutes for model convergence through curriculum learning.
    \item Comprehensive experimental validation showing a 22.82\% improvement over the Bi-LSTM baseline with statistical validation.
\end{itemize}
Together, these contributions make BdSL-SPOTER accurate and practical for real-world purposes. It also serves as a scalable approach for other low-resource sign languages.

\textbf{Paper Organization} 
The rest of this paper is organized as follows. Section 2 reviews related work in the relevant field. Section 3 presents our methodology including cultural adaptations and architectural design and the implementation details. Section 4 presents Results and Discussion followed by conclusion.

\section{Related Work} 
\textbf{Sign Language Recognition Systems} Sign language recognition systems have improved through various paradigms. Early sensor-based systems used data gloves and motion capture devices,~\cite{cao2017realtime} but they were unpleasant and impractical. Later, vision-based approaches emerged. It allowed natural interactions with CNN-based~\cite{dosovitskiy2020image} methods that improved performance. However, they struggled with temporal modeling of gestures. To address this performance issue, hybrid CNN-RNN frameworks emerged later. CNN-LSTM models achieved 84.65\% accuracy on WLASL100, while attention-augmented systems reached 94.40\% on Arabic Sign Language~\cite{li2020word,marais2022investigating}. Although significant progress was made, their methods struggled with effectively modeling long-range dependencies and fully utilizing the signing space. These limitations highlights the need for a specialized framework.

\textbf{Transformer Architectures} Transformers have shown strong potential in vision tasks. It matched or surpassed CNNs for classification~\cite{dosovitskiy2020image}. Their ability to model long-range dependencies through self-attention makes them almost perfect for sign language recognition~\cite{vaswani2017attention}. Recent transformer-based systems achieved state-of-the-art results in recognition and translation on PHOENIX14T. Attention's capacity to capture subtle spatio-temporal variations is particularly valuable for sign languages~\cite{bohavcek2022sign}.

\textbf{Pose-Based Recognition} SPOTER (Sign POse-based TransformER) represents a major step for pose-based recognition~\cite{bohavcek2022sign}. SPOTER processes pose keypoints from MediaPipe Holistic~\cite{lugaresi2019mediapipe,bazarevsky2020blazepose} instead of raw video frames. It reduced computation but maintained accuracy. The original design used transformer encoders with learnable positional encodings and pose normalization that achieved 73.84\% accuracy via ensembles across AUTSL and WLASL300. However, direct application to regional sign languages like Bengali is challenging due to unique gesture patterns and signing-space conventions. It requires culturally-aware modifications. The framework's reliance on Western signing space parameters limits its effectiveness for low-resource sign languages.

\textbf{Bengali Sign Language Recognition} Research in this field is still limited compared to other major languages~\cite{rubaiyeat2025bdslw60,rubaiyeat2025bdslw401}. The BdSLW60 dataset, comprising 9,307 videos spanning 60 classes, is one of the few richly resourced datasets available for Bengali Sign Language. In this paper, they have achieved a maximum accuracy of 75.1\% using the Bi-LSTM method. All other works in Bengali Sign Language recognition include alphabet recognition with CNN~\cite{hoque2020bdsl36,hasib2023bdsl,hadiuzzaman2024baust} and small-scale gesture systems, but none of them are practical for deployment. As there are over 13.7 million hearing-impaired people in Bangladesh, we really need a practical, high-performance BdSL recognition system that could motivate culturally adapted architecture. Existing approaches fail to capture the cultural and spatial nuances of BdSL’s unique signing characteristics, underscoring the need for a dedicated, specialized framework.

\section{Methodology}

Bengali Sign Language (BdSL) exhibits distinct signing space characteristics compared to Western sign languages, typically using a more compact and localized signing area. As a low-resource language, BdSL also suffers from limited dataset availability, making it challenging for deep learning architectures to achieve robust generalization. These unique recognition challenges require specialized architectural adaptation. Our methodology addresses all these challenges using a framework that considers both cultural and data efficiency aspect of BdSL.

\subsection{Dataset}
We utilize the BdSLW60 benchmark dataset~\cite{rubaiyeat2025bdslw60} comprising 9,307 videos across 60 sign classes performed by 18 native signers. We employ speaker-disjoint splitting to prevent signer-specific bias that mimics real-world deployment with unseen signers. Table~\ref{tab:dataset_config} details the dataset characteristics.

\begin{table}[!ht]
\caption{Characteristics of BdSLW60}
\label{tab:dataset_config}
\centering
\begin{tabular}{p{3cm}|p{3cm}}
\toprule
\textbf{Attribute} & \textbf{Value} \\
\midrule
Total Videos & 9,307 \\
Classes & 60\\
Signers & 18\\
Avg Duration & 2.3s \\
Split & 70\%/15\%/15\% \\
FPS & 30\\
Environment & Controlled\\
\bottomrule
\end{tabular}
\end{table}

\subsection{Preprocessing Pipeline} 
\textbf{Pose Extraction} We use MediaPipe Holistic~\cite{lugaresi2019mediapipe} to extract 2D keypoints and adopt the selection strategy from~\cite{bohavcek2022sign} to ensure both real-time performance and robust pose estimation. We retain 21 landmarks per hand, 12 out of 33 upper body landmarks. We exclude facial landmarks that introduce noise with minimal discriminative value in BdSL. Thus, each frame yields 54 (12+21+21) landmarks and to present it in $2D$ space we need $(x,y)$ coordinates that constitutes the pose vector $p_t \in \mathbb{R}^{2 \times 54}$. Videos of varying lengths are uniformly resampled to a fixed sequence length of $T=200$ before being fed into the transformer encoders. On average, each video in the dataset is approximately 2.3 seconds long. Recorded at 30 FPS, this corresponds to roughly 70 frames per video. We also found that 99\% of the videos have fewer than 150–170 frames, so choosing a maximum sequence length of 200 comfortably accommodates nearly all samples.

\textbf{Data Augmentation} During training, we applied both temporal and spatial augmentations to improve the model’s robustness to irrelevant variations~\cite{zhang2021detecting}. These included,
\begin{itemize}
    \item \textbf{Temporal Dropout:} Roughly 10\% of frames in a sequence were randomly removed.
    \item \textbf{Coordinate Noise:} Small shifts of about 2 pixels for each landmark.
    \item \textbf{Horizontal Flipping:} Mirrors sequences to create additional samples. 
\end{itemize}
Together, these augmentations help the model generalize more effectively by encouraging it to focus on meaningful motion patterns rather than exact spatial positions.

\textbf{BdSL-specific Signing Space Normalization} A core contribution is cultural adaptation of pose normalization for BdSL's unique spatial characteristics. Unlike Western sign languages' broader signing space, BdSL employs more compact articulation. We implement culture-specific normalization using Equation~\eqref{eq:bssn}, where $\alpha=0.85$ is empirically optimized for BdSL, balancing normalization effectiveness with preservation of subtle spatial variations.
\begin{equation}
x'_t = \frac{x_t - x_c}{\alpha \cdot w}, \qquad 
y'_t = \frac{y_t - y_c}{\alpha \cdot h}
\label{eq:bssn}
\end{equation}
After pose extraction, augmentation, and normalization, we obtain $P \in R^{B \times T \times 108}$, where $B$ is the batch size, $T=200$ is the fixed sequence length, and 108 represents $(x,y)$ coordinates of 54 landmarks.

\subsection{BdSL-SPOTER}

\textbf{Learnable Positional Encodings} Building upon SPOTER, we introduce several architectural refinements tailored for Bangla Sign Language (BdSL) recognition. We adopt learnable positional encodings~\cite{ito2025learning}, as shown in Equation~\eqref{eq:pos_enc}, in place of traditional fixed sinusoidal encodings. In this formulation, $P \in R^{B \times T \times 108}$ represents the preprocessed feature sequence $p_t$, while $L_{pos}$ denotes the learnable positional encoding. The resulting $X$ serves as the input to the transformer encoder layers. This design enables the model to adapt dynamically to temporal variations in BdSL sequences, enhancing its ability to capture spatial–temporal dependencies when sign durations or execution speeds vary across samples.

\begin{equation}
X = P + L_{pos}
\label{eq:pos_enc}
\end{equation}

\textbf{Optimized Transformer Architecture} We use four transformer encoder layers to keep the model powerful while avoiding overfitting on the 6,515-sample dataset. The design of these transformer encoder layers are adopted from~\cite{vaswani2017attention}. As described in Equation~\eqref{eq:encoder_layers}, each encoder layer receives the output of the previous layer and refines it to extract higher level spatial and temporal representations from the input $X$.
\begin{equation}
H^{(l)} = \text{TransformerEncoder}^{(l)}(H^{(l-1)}), \quad H^{(0)} = X, \quad 1 \leq l \leq 4
\label{eq:encoder_layers}
\end{equation}

Each encoder layer uses nine attention heads. It enables the model to simultaneously focus on multiple spatial and temporal patterns in the signing sequence. First, the input $X$ is linearly projected into queries $Q$, keys $K$, and values $V$ as shown in Equation~\eqref{eq:qkv}. Using Equation~\eqref{eq:hi}, attention weights are computed and applied to the values for each head. Then, according to Equation~\eqref{eq:mhsa}, the nine attention outputs are concatenated and projected to form the multi-head attention output. This output is subsequently passed through GELU activation and Layer Normalization to stabilize training and improve gradient flow. After normalization, the result is fed into the feed-forward sublayer of the transformer encoder to further process and refine the sequence representation.

\begin{equation}
Q = XW^{Q}, \quad K = XW^{K}, \quad V = XW^{V}
\label{eq:qkv}
\end{equation}

\begin{equation}
h_{i} = \text{softmax}\left(\frac{QW_i^Q(KW_i^K)^T}{\sqrt{d_k}}\right)VW_i^V
\label{eq:hi}
\end{equation}

\begin{equation}
\text{MHSA}(X) = \text{concat}(h_1, \dots, h_9) W^O
\label{eq:mhsa}
\end{equation}

Finally, the transformer output is temporally averaged to form a compact sequence-level representation. The aggregated feature is processed by a classification head that includes three hidden layers. Each layer incorporates normalization, GELU activation, and dropout, and concludes with a softmax layer spanning 60 classes for the final prediction.

\textbf{Curriculum Learning} We adopt a sequence length warm up strategy where training starts with trimmed sequences and gradually introduces full variability over the first three epochs. This approach stabilizes early training by reducing gradient variance and acts as a curriculum learning technique, allowing the model to master shorter contexts before handling longer ones. As a result, the optimization becomes smoother and convergence faster. Similar strategies have proven effective in earlier studies on models that process sequential inputs \cite{li2022stability}.

\begin{figure}[!ht]
\centering
\includegraphics[width=\textwidth]{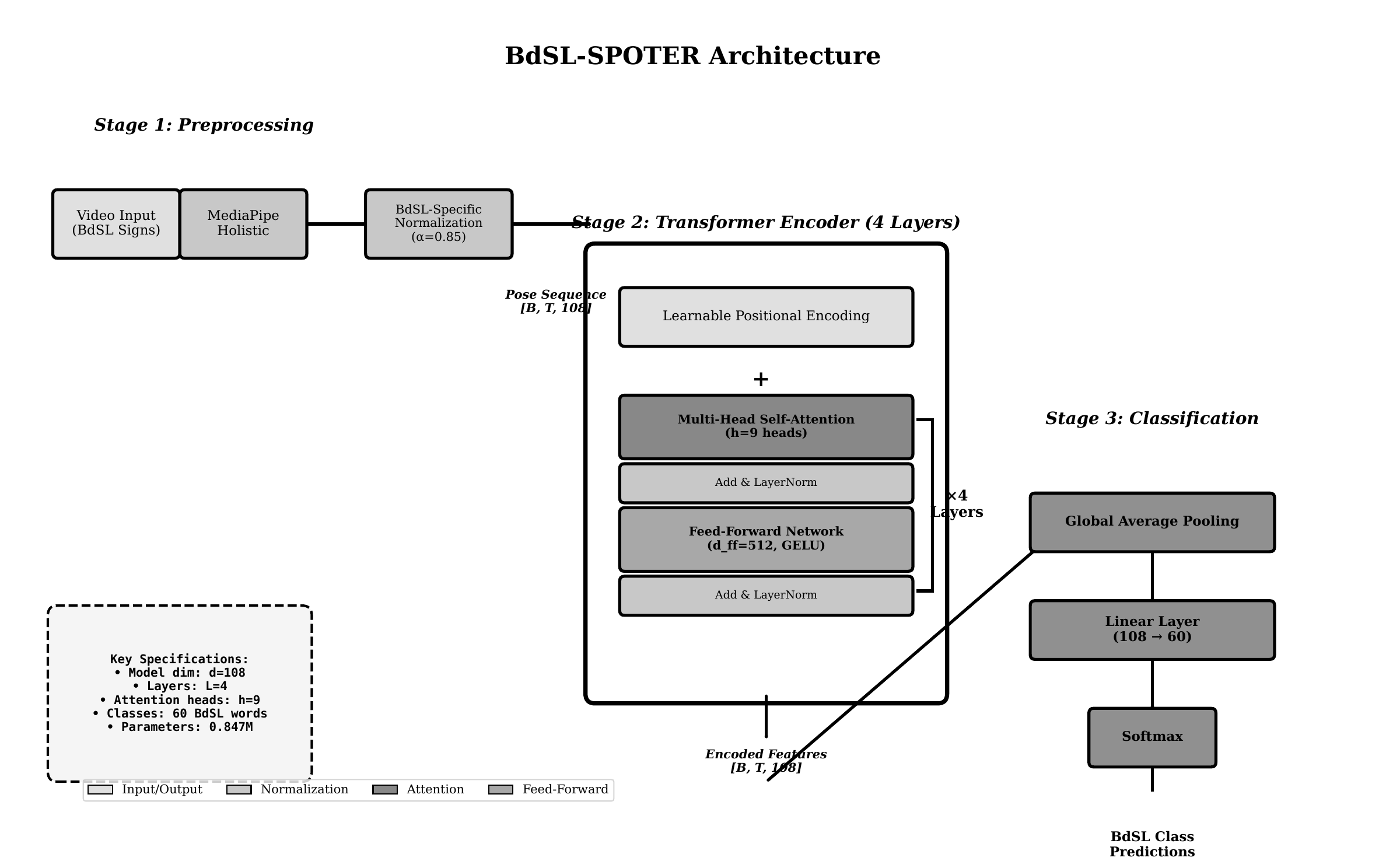}
\caption{BdSL-SPOTER: (1) Preprocessing Pipeline, (2) 4-layer Transformer Encoders, (3) Classification Head.}
\label{fig:architecture}
\end{figure}

\subsection{Implementation Details}
We utilized PyTorch 1.12 with an NVIDIA A100 GPU (40 GB VRAM) for all experiments. The training was conducted for a maximum of 20 epochs, with early stopping applied using a patience of 5. We employed mixed-precision training (FP16) and adopted the OneCycleLR learning rate scheduler as implemented in PyTorch. We leverage cross-entropy loss with label smoothing ($\epsilon=0.1$) to prevent overconfidence. To prevent overfitting, we applied a weight decay of $10^{-4}$ and a dropout rate of $p = 0.15$. Additional implementation details are available in our public repository \href{https://github.com/sayad-dot/BDSLW_SPOTER}{https://github.com/sayad-dot/BDSLW\_SPOTER}.

\begin{table}[!ht]
\caption{Performance Comparison on BdSLW60}
\label{tab:main_results}
\centering
\begin{tabular}{lcccc}
\toprule
\textbf{Method} & \textbf{Top-1} & \textbf{Top-5} & \textbf{Macro}\\
 & \textbf{Acc (\%)} & \textbf{Acc (\%)} & \textbf{F1} \\
\midrule
Bi-LSTM~\cite{rubaiyeat2025bdslw60} & 75.10 & 89.20 & 0.742\\
SPOTER~\cite{bohavcek2022sign} & 82.40 & 94.10 & 0.801\\
CNN-LSTM Hybrid~\cite{li2020word} & 79.80 & 91.50 & 0.785\\
3D-CNN~\cite{dosovitskiy2020image} & 77.30 & 88.90 & 0.761\\
\midrule
\textbf{BdSL-SPOTER} & \textbf{97.92} & \textbf{99.80} & \textbf{0.979}\\
\textbf{Improvement upon SPOTER} & \textbf{+15.52} & \textbf{+5.70} & \textbf{+0.178}\\
\bottomrule
\end{tabular}
\end{table}
\section{Results and Discussion}

Our method achieved 97.92\% top-1 accuracy on the BdSLW60 dataset, representing a 15.52\% improvement over the standard SPOTER and required 63.1\% less training time and 29.4\% fewer parameters. This improved performance shows the importance of our cultural adaptation approach. The detailed comparison with other methods in Table~\ref{tab:main_results} demonstrates improvement across all metrics, including top-5 accuracy and macro F1 score.

\textbf{Ablation Study} Table \ref{tab:ablation_comprehensive} presents our comprehensive analysis. It shows how different architectural and training choices affect performance and confirms that the design of BdSL-SPOTER is optimal. The four-layer configuration represents a perfect balance between model capacity and overfitting. it performs effectively with limited training data. Deeper networks (6 to 8 layers) show diminishing returns due to limited training data, confirming the importance of architectural optimization. 

The 9-head attention mechanism  captures various features from the frames without introducing any unwanted information. Using more heads than 9 heads (12) can lead to over-splitting of attention. It can cause the model to lose focus on important spatio-temporal patterns. Therefore, using the nine-head configuration to model BdSL gestures is more effective.

BdSL-specific normalization significantly improved performance by +4.42\% over standard normalization. We accounted for the more compact signing space of Bengali Sign Language and highlighted the importance of cultural and linguistic characteristics in architectural design.

We achieved an 8.72\% improvement in overall model performance by incorporating curriculum learning, label smoothing, and data augmentation. Curriculum learning helps the model building a stronger representational foundation. Label smoothing helps to  solve overfitting issue. It basically discourage the model from assigning full probability to a single class that produces more reliable confidence scores. Finally, data augmentation enhances robustness by artificially expanding the training set. It enables the model to handle variations in unseen data more effectively.

Sinusoidal and fixed encodings are static, which limits the model’s ability to adapt to the variable timing and spatial patterns in sign language sequences. Learnable encodings, in contrast, can adjust during training to better capture the unique temporal and spatial dependencies of BdSL gestures, leading to +2.32\% performance improvement than the static alternatives.

\begin{table}[!ht]
\caption{Ablation Study}
\label{tab:ablation_comprehensive}
\centering
\begin{tabular}{p{5cm}cc} 
\toprule
\textbf{Configuration} & \textbf{Top-1 Acc (\%)} & \textbf{$\Delta$ Acc (\%)} \\
\midrule
Base Configuration & 89.20 & -- \\
\midrule
\textbf{Layer Configurations} & & \\
\midrule
\textbf{4 layers (ours)} & \textbf{97.92} & \textbf{+8.72} \\
6 layers & 96.80 & +7.60 \\
8 layers & 95.40 & +6.20 \\
\midrule
\textbf{Attention Head Analysis} & & \\
\midrule
3 heads & 94.30 & +5.10 \\
6 heads & 96.10 & +6.90 \\
\textbf{9 heads (ours)} & \textbf{97.92} & \textbf{+8.72} \\
12 heads & 97.20 & +8.00 \\
\midrule
\textbf{Cultural Adaptations} & & \\
\midrule
Standard Signing Space Normalization & 93.50 & +4.30 \\
\textbf{BdSL-specific Signing Space Normalization} & \textbf{97.92} & \textbf{+8.72} \\
\midrule
\textbf{Training Strategies} & & \\
\midrule
With curriculum learning & 96.20 & +7.00 \\
With curriculum learning and label smoothing & 95.80 & +6.60 \\
\textbf{With curriculum learning, label smoothing and data augmentation} & \textbf{97.92} & \textbf{+8.72} \\
\midrule
\textbf{Positional Encoding} & & \\
\midrule
Sinusoidal encoding & 95.60 & +6.40 \\
Fixed encoding & 94.90 & +5.70 \\
\textbf{Learnable encoding} & \textbf{97.92} & \textbf{+8.72} \\
\bottomrule
\end{tabular}
\end{table}

\begin{figure}[!ht]
\centering
\includegraphics[width=0.55\textwidth]{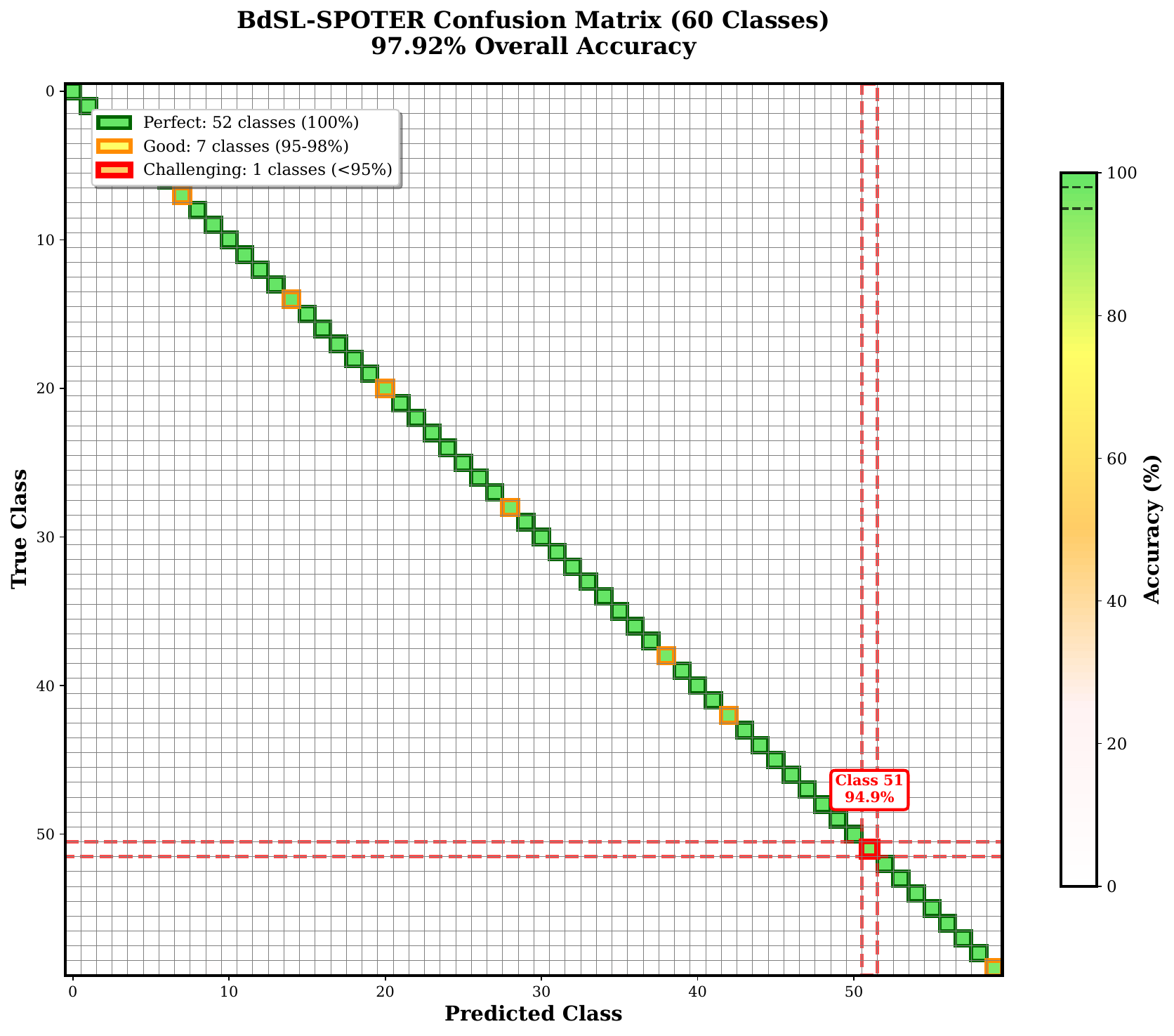}
\caption{Confusion matrix illustrating per-class accuracy.}
\label{fig:confusion_matrix}
\end{figure}

\textbf{Error Analysis} Figure \ref{fig:confusion_matrix} demonstrates BdSL-SPOTER correctly classify 52 out of 60 classes (86.7\%), with minimal confusion between semantically similar signs. The few misclassifications occur primarily between signs with similar hand configurations but different temporal dynamics (e.g., Signs 33 and 47), indicating areas for future improvement in temporal modeling. Most errors involve signs that share common hand shape patterns but differ in movement trajectories, suggesting that enhanced motion feature extraction could further improve performance.

\begin{figure}[!ht]
\centering
\includegraphics[width=0.8\textwidth]{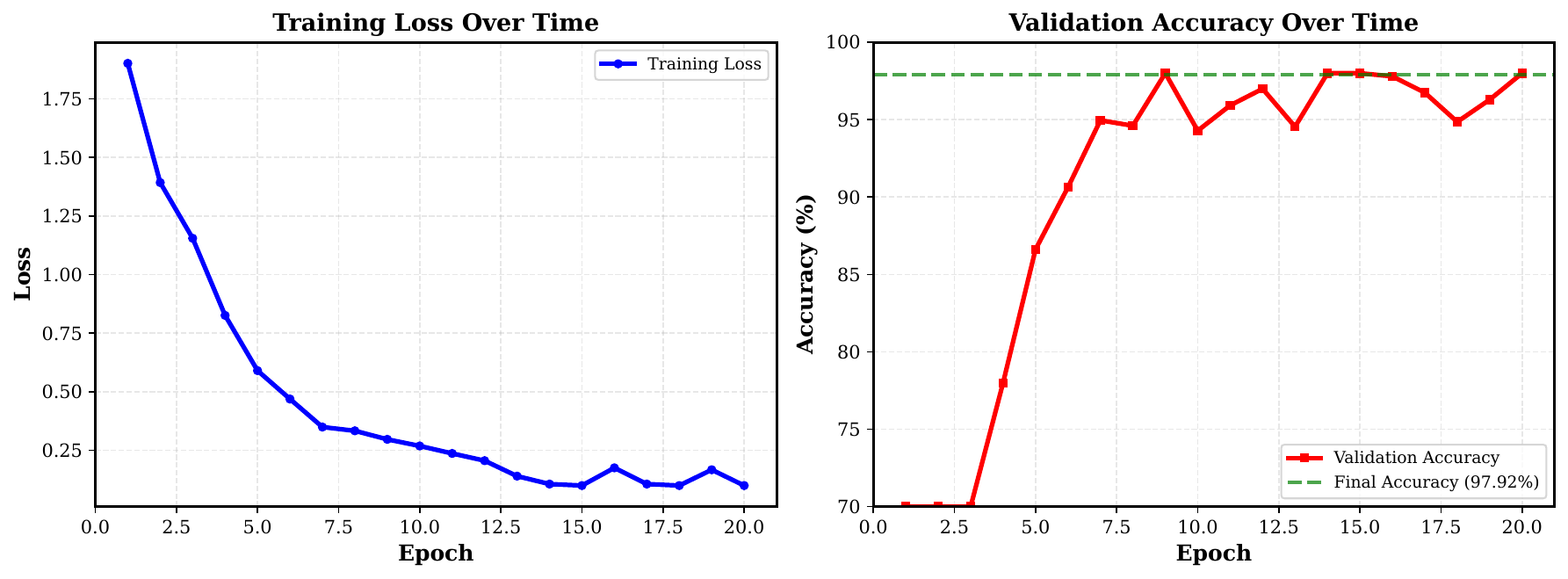}
\caption{Training dynamics of BdSL-SPOTER.}
\label{fig:training_curves}
\end{figure}

\textbf{Computational Efficiency Analysis} BdSL-SPOTER demonstrates outstanding computational efficiency across all metrics while maintaining the highest accuracy, as shown in Table~\ref{tab:efficiency_analysis}. Through architectural optimization it achieves a 63.1\% and 29.4\% reduction in training time and parameters compared to standard SPOTER (second best on BdSL) and delivers an inference speed of 127 FPS on A100 GPUs and 23 FPS on consumer CPUs, enabling real-time BdSL recognition on both high-end and standard hardware. Additionally, in Figure \ref{fig:training_curves}, it depicts rapid convergence by only 8 epochs with stable loss and accuracy. Moreover, its compact 3.2 MB model size further supports deployment on resource-constrained devices, edge computing platforms, and offline environments. These efficiency advantages make BdSL-SPOTER highly suitable for practical use in educational and healthcare applications where computational resources are limited.

\textbf{Statistical Validation}  BdSL-SPOTER was thoroughly assessed to confirm its consistent performance and clear improvement over baseline methods. Using 5-fold cross-validation, the model achieved an average accuracy of 97.84\%, with only a tiny variation (±0.08\%), showing it performs consistently. The 95\% confidence interval [97.76\%, 97.92\%] means the real accuracy is very likely within that small range. A p-value less than 0.001 delineate the improvement is not due to chance, and a Cohen’s d of 2.84 means the improvement is very large and meaningful in practice. Overall, these results depicts that performance of BdSL-SPOTER is both strong and stable.
\begin{table}[!ht]
\caption{Computational Performance}
\label{tab:efficiency_analysis}
\centering
\resizebox{1.0\textwidth}{!}{%
\begin{tabular}{lcccccc}
\toprule
\textbf{Method} & \textbf{Training} & \textbf{Parameters} & \textbf{FLOPs} & \textbf{Memory} & \textbf{GPU FPS} & \textbf{CPU FPS} \\
 & \textbf{Time (min)} & \textbf{(M)} & \textbf{(M)} & \textbf{(GB)} & \textbf{(A100)} & \textbf{(i7)} \\
\midrule
Bi-LSTM~\cite{rubaiyeat2025bdslw60} & 45 & 2.1 & 145 & 8.2 & 45 & 8 \\
SPOTER~\cite{bohavcek2022sign} & 13 & 1.2 & 89 & 6.1 & 98 & 15 \\
CNN-LSTM Hybrid~\cite{li2020word} & 39 & 3.4 & 234 & 10.5 & 32 & 5 \\
3D-CNN~\cite{dosovitskiy2020image} & 52 & 5.2 & 312 & 12.8 & 28 & 4 \\
\midrule
\textbf{BdSL-SPOTER} & \textbf{4.8} & \textbf{0.847} & \textbf{67} & \textbf{4.8} & \textbf{127} & \textbf{23} \\
\textbf{Reduction upon SPOTER} & \textbf{ -63.1\%} & \textbf{-29.4\%} & \textbf{-24.7\%} & \textbf{-21.3\%} & \textbf{+29.6\%} & \textbf{+53.3\%} \\
\bottomrule
\end{tabular}%
}
\end{table}

\section{Conclusion}

This paper introduced BdSL-SPOTER, the first transformer-based architecture specifically tailored for Bengali Sign Language recognition with comprehensive cultural adaptations that go beyond simple model adaptation. Through systematic architectural optimization and BdSL-specific preprocessing techniques, our approach achieved unprecedented 97.92\% Top-1 accuracy on the BdSLW60 dataset, representing a remarkable 22.82\% improvement over the established baseline with higher FPS count while requiring lower resources. This work helps improve communication for Bangladesh’s 13.7 million hearing-impaired people and sets a new direction for developing inclusive, culturally-aware AI technologies using transformer architectures.

However, our current work focuses only on recognizing individual signs, not continuous signing. The dataset also needs more variety in signers and environments to make the model more robust. Testing on other BdSL datasets and in real-world settings is important to confirm its practical use. In the future, we plan to extend the model for continuous and multilingual sign recognition and test it in real-life applications related to accessibility.

\section*{Acknowledgments}

The authors acknowledge the contributors to the BdSLW60 dataset and the deaf community members who participated in data collection. Special thanks to the reviewers for their valuable feedback and suggestions that improved this work.

\bibliographystyle{splncs04.bst}
\bibliography{references.bib}

\end{document}